\documentclass[12pt]{article}

\usepackage{scicite}

\usepackage{times}

\usepackage[version=3]{mhchem} 
\usepackage{graphicx}
\usepackage{subcaption}
\usepackage{longtable} 
\usepackage{notes2bib}
\usepackage{url}
\usepackage{array}
\usepackage{hyperref}
\usepackage{multirow}
\usepackage{marginnote}
\usepackage{bm}
\usepackage{physics}
\usepackage{braket}
\usepackage[ruled,vlined]{algorithm2e}

\usepackage{xcolor}

\newenvironment{sciabstract}{%
\begin{quote} \bf}
{\end{quote}}

\title{Evolving symbolic density functionals}

\author{He Ma,$^{1}$ Arunachalam Narayanaswamy,$^{1}$ Patrick Riley,$^{1}$ Li Li,$^{1\ast}$\\
\normalsize{$^{1}$Google Research, Mountain View, CA 94043, USA}\\
\normalsize{$^\ast$To whom correspondence should be addressed; E-mail:  leeley.lili@gmail.com}
}

\date{}

\begin{document} 


\baselineskip24pt


\maketitle


\begin{sciabstract}
Systematic development of accurate density functionals has been a decades-long challenge for scientists. Despite the emerging application of machine learning (ML) in approximating functionals, the resulting ML functionals usually contain more than tens of thousands parameters, which makes a huge gap in the formulation with the conventional human-designed symbolic functionals. We propose a new framework, Symbolic Functional Evolutionary Search (SyFES), that automatically constructs accurate functionals in the symbolic form, which is more explainable to humans, cheaper to evaluate, and easier to integrate to existing density functional theory codes than other ML functionals. We first show that without prior knowledge, SyFES reconstructed a known functional from scratch. We then demonstrate that evolving from an existing functional $\omega$B97M-V, SyFES found a new functional, GAS22 (Google Accelerated Science 22), that performs better for the majority of molecular types in the test set of Main Group Chemistry Database (MGCDB84). Our framework opens a new direction in leveraging computing power for the systematic development of symbolic density functionals.
\end{sciabstract}


\section*{Introduction}

Quantum mechanical simulations of molecules and materials are playing an increasingly important role in chemistry, physics and materials sciences. Density functional theory (DFT)~\cite{hohenberg1964inhomogeneous} has been one of the most successful methods for determining electronic structures of molecules and materials from first principles~\cite{martin2020electronic,RevModPhys.87.897}, and has been widely used for the design and characterization of novel drugs~\cite{mardirossian2020novel}, catalysts~\cite{norskov2011density} and functional materials~\cite{galli2020long}.
Most DFT calculations performed today adopt the Kohn-Sham (KS) scheme~\cite{kohn1965self}. KS-DFT maps the challenging problem of solving the many-body Schrodinger equation of interacting electrons into the solution of one-body Kohn-Sham equations, with the complicated many-body effect treated with the exchange-correlation (XC) functional. This mapping is in principle exact. However, since the exact form of the XC functional is unknown, approximate forms are required in practice and the accuracy of results are limited by the quality of such approximations.

The development of accurate XC functionals has been an important subject for decades~\cite{becke2014perspective,yu2016perspective,li2020recent}. To date, researchers have proposed more than 200 different XC functionals~\cite{mardirossian2017thirty}. Most functionals used today contain a few to a few dozens of empirical parameters, which are usually determined by fitting to datasets of molecular or materials properties. Many widely-used XC functionals, such as those developed by Head-Gordon and coworkers~\cite{chai2008systematic,mardirossian2014omegab97x,mardirossian2015mapping,mardirossian2016omega,mardirossian2018survival} and the well known Minnesota functionals~\cite{zhao2011applications,peverati2011improving,yu2016mn15,wang2020m06}, are constructed by taking linear combinations of expressions inspired by existing functional forms (e.g. the B97 functional~\cite{becke1997density}), where the linear coefficients and other empirical parameters are fit to databases such as the Main Group Chemistry Database (MGCDB84)~\cite{mardirossian2016omega} and Minnesota Database~\cite{peverati2014quest}.

Despite great efforts, it is generally considered difficult to develop more accurate functionals than existing ones in a systematic manner. In the past decade, researchers have devoted great efforts to approximate functionals using machine learning (ML)~\cite{KLMB21}. One direction is to accelerate DFT with accurate kinetic energy functional approximation or bypass the Kohn-Sham equations using kernel ridge regression~\cite{snyder2012finding,li2016understanding,li2016pure,brockherde2017bypassing,bogojeski2020quantum} and neural networks~\cite{seino2018semi,ryczko2019deep,fujinami2020orbital,meyer2020machine}. 
The other direction is to solve the decades-long challenge -- fundamentally improving the accuracy of DFT with better XC functionals.
Various ML techniques have been applied, e.g. Bayesian error estimation~\cite{wellendorff2012density,lundgaard2016mbeef}, linear regression with subset selection procedure~\cite{mardirossian2014omegab97x,mardirossian2015mapping,mardirossian2016omega,mardirossian2018survival}, genetic algorithm~\cite{gastegger2019exploring} and Bayesian optimization~\cite{vargas2020bayesian}. In these works, the functional forms are usually chosen a priori or selected from a relatively rigid space of functional forms. Furthermore, many parameters in these forms are linear in nature, which has the advantage of being easily optimizable but limits the expressive power of the functional form.
In contrast, neural networks are able to approximate any continuous function~\cite{hornik1991approximation} and thus are flexible approximators to parameterize XC functionals. 
Such neural networks can be trained with self-consistent DFT calculations via differentiable programming~\cite{li2021kohn,kasim2021learning,dick2021using}, simulated annealing~\cite{nagai2020completing,PhysRevResearch.4.013106}, or on converged DFT or beyond DFT calculations~\cite{chen2020deepks,dick2020machine,dm21}.
Although neural networks XC functionals with more than tens of thousands parameters can achieve high accuracy for particular systems, they are less explainable to humans, expensive to execute, and difficult to integrate to existing DFT codes compared to conventional human-designed symbolic forms.

\begin{figure*}[!h]
  \centering
  \includegraphics[width=7in]{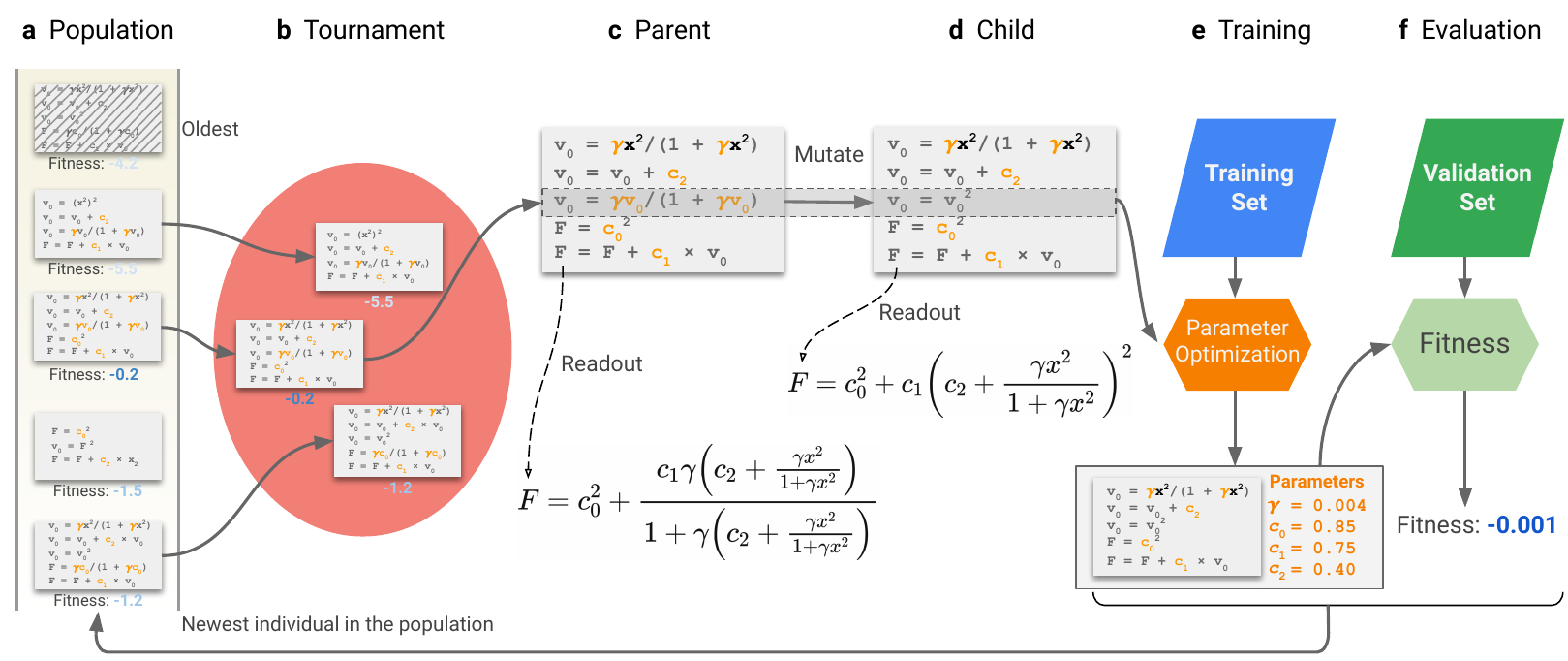}
  \caption{{\bf Workflow of the Symbolic Functional Evolutionary Search (SyFES) framework.} {\bf a} A population of symbolic density functionals is iteratively evolved using the regularized evolution algorithm. Each individual in the population represents the enhancement factors in a density functional. The performance of a density functional is characterized by its fitness, defined as the negative of validation error in kcal/mol. {\bf b} In each iteration, a tournament selection is performed on a subset of individuals sampled from the population. {\bf c} The highest fitness individual in the tournament is selected to be the parent. {\bf d} Then a child functional is generated by randomly picking one of the enhancement factor and mutating one of its instructions. {\bf e} The child functional is then trained on the training set using the covariance matrix adaptation evolution strategy (CMA-ES) method to determine its parameters. {\bf f} Finally, the fitness of child functional is evaluated on the validation set, and then added to the population. After the population size exceeds a limit, the oldest individual in the population is removed. For simplicity of visualization, a density functional is represented with only one enhancement factor. The general form of density functional considered in this work contains 3 enhancement factors ($F_\text{x}$, $F_\text{c-ss}$, $F_\text{c-os}$), and in each evolution step one enhancement factor is randomly chosen for mutation.}
  \label{workflow}
\end{figure*}

In this paper, we propose a new approach to develop more accurate functionals -- searching XC functionals in a large, nonlinear, symbolic functional space based on the concept of {\it symbolic regression}. Unlike most ML methods where models are formulated numerically, symbolic regression outputs the resulting {\it model} in the symbolic form. Recently, there is emerging interest in development of symbolic regression methods for physical science problems~\cite{schmidt2009distilling,doi:10.1126/sciadv.aay2631,ouyang2018sisso,doi:10.1126/sciadv.aav0693,ngmcts,cranmer2020,hernandez2019fast,kabliman2021application,wang2019symbolic,vaddireddy2020feature,kharkov2021discovering,sun2022symbolic}.
We illustrate our framework in Fig.~\ref{workflow} and denote it as Symbolic Functional Evolutionary Search (SyFES).
One key component of SyFES is a symbolic representation of XC functionals based on elementary mathematical instructions and building blocks of existing functionals. The symbolic representation mimics the execution of XC functionals by computer programs, and we demonstrate this representation enables efficient search of functional forms. Then, using a genetic algorithm called regularized evolution~\cite{real2019regularized}, we demonstrate that simple functionals such as the B97 exchange functional can be obtained from scratch, and that more accurate functionals can be obtained by evolving from existing functionals. In particular, from a set of regularized evolution starting from the $\omega$B97M-V functional~\cite{mardirossian2016omega}, we discovered a novel functional form GAS22 (Google Accelerated Science 22) with lower test error on the MGCDB84 dataset, which is the dataset that the $\omega$B97M-V functional is originally trained on. We further demonstrate that GAS22 exhibits good numerical stability for self-consistent calculations.

\section*{Results}

\noindent {\bf Representation of exchange-correlation functionals}

In KS-DFT, the total energy $E_\text{tot}$ for a system of interacting electrons is represented as a functional of electron density $\rho$:
\begin{equation}
    E_\text{tot}[\rho] = T_s[\rho] + E_\text{ext}[\rho] + E_\text{H}[\rho] + E_\text{xc}[\rho]
\label{etot}
\end{equation}
where $T_s[\rho]$, $E_\text{ext}[\rho]$, $E_\text{H}[\rho]$, $E_\text{xc}[\rho]$ denotes the Kohn-Sham kinetic energy, external energy, Hartree energy and the XC energy, respectively. Only the exact form for $E_\text{xc}[\rho]$ is unknown and approximate forms must be used in practice. Most existing approximate forms for $E_\text{xc}[\rho]$ contains a semilocal part $E_\text{xc}^\text{sl}$, and many modern functional forms also contains a nonlocal part $E_\text{xc}^\text{nl}$. The overall $E_\text{xc}$ can thus be written as 
\begin{equation}
    E_\text{xc}[\rho(\bm{r})] = E_\text{xc}^\text{sl}[\rho(\bm{r})] + E_\text{xc}^\text{nl}[\rho(\bm{r})].
\label{exc}
\end{equation}
The semilocal part $E_\text{xc}^\text{sl}$ can generally be written as an integral of XC energy density $e_\text{xc}$ over real space coordinate $\bm{r}$, with $e_\text{xc}$ being a function of $\rho$ and its various orders of derivatives. The semilocal part is generally the energetically dominant component of $E_\text{xc}$, and is the main focus in this work. The nonlocal part $E_\text{xc}^\text{nl}$ is usually introduced to address certain interactions that are difficult to capture by the semilocal part, e.g. dispersion interaction.

We fix the nonlocal part $E_\text{xc}^\text{nl}$ being identical to that of the $\omega$B97M-V functional and search for better semilocal part $E_\text{xc}^\text{sl}$.
For simplicity we present formulas for spin-unpolarized systems in the main text, and in Supplementary Materials (SM) we present the general formalism for spin-polarized systems.
In the semilocal part, 
\begin{equation}
    E_\text{xc}^\text{sl}[\rho] = \int \left( e^\text{LDA}_\text{x-sr}[\rho] F_\text{x} + e^\text{LDA}_\text{c-ss}[\rho] F_\text{c-ss} + e^\text{LDA}_\text{c-os}[\rho] F_\text{c-os} \right) \text{d}\bm{r},
\label{excsl}
\end{equation}
terms $e^\text{LDA}_\text{x-rs}[\rho]$, $e^\text{LDA}_\text{c-ss}[\rho]$, and $e^\text{LDA}_\text{c-os}[\rho]$ denote short-range exchange, same-spin correlation and opposite-spin correlation energy densities within the local density approximation (LDA)~\cite{perdew1992accurate}, respectively. Their explicit forms are known (see SM for the expression of $e^\text{LDA}_\text{x-sr}$ and Ref.~\citen{stoll1978theor} for expressions of $e^\text{LDA}_\text{c-ss}$ and $e^\text{LDA}_\text{c-os}$). $F_\text{x}$, $F_\text{c-ss}$ and $F_\text{c-os}$ are the exchange, same-spin correlation and opposite-spin enhancement factors that represent corrections over LDA energy densities. The enhancement factors can depend on density gradient for GGA functionals and additionally on kinetic energy density for meta-GGA functionals. This form has been adopted by many functionals, most notably the B97 functional~\cite{becke1997density} and many B97-inspired functionals~\cite{zhao2011applications,peverati2011improving,yu2016mn15,wang2020m06,chai2008systematic,mardirossian2014omegab97x,mardirossian2015mapping,mardirossian2016omega,mardirossian2018survival}.
The functional form in equation~(\ref{exc}) is then entirely determined by enhancement factors ($F_\text{x}$, $F_\text{c-ss}$, and $F_\text{c-os}$). Thus we will use the term XC functional and enhancement factors interchangeably in the following discussions. We note that although we selected a particular form of $E_\text{xc}^\text{sl}$ and $E_\text{xc}^\text{nl}$, the SyFES framework is general and can be applied to other forms.

\hspace{1cm}

\noindent {\it Symbolic representation for machine}

Human representation of the XC functionals is symbolic. For machines to automatically search the symbolic forms of the XC functionals, we need a machine representation that can both be translated to the human representation and be easily modified by search algorithms.
In this work, we represent the form of XC functionals as the execution of a set of mathematical \textit{instructions}.
This representation is inspired by recent progress in the field of automated ML (AutoML)~\cite{real2020automl,coreyes2021evolving}, also known as a form of linear genetic programming~\cite{brameier2007linear}. For instance, Ref.~\citen{real2020automl} showed that by representing a computer program as a sequence of instructions, a regularized evolution algorithm~\cite{real2019regularized} can learn to construct complicated computer programs from a set of instructions for various ML tasks. Notably, it can automatically rediscover many ML techniques that researchers developed and used in the past decades.

\begin{figure*}[!h]
  \centering
  \includegraphics[width=3.5in]{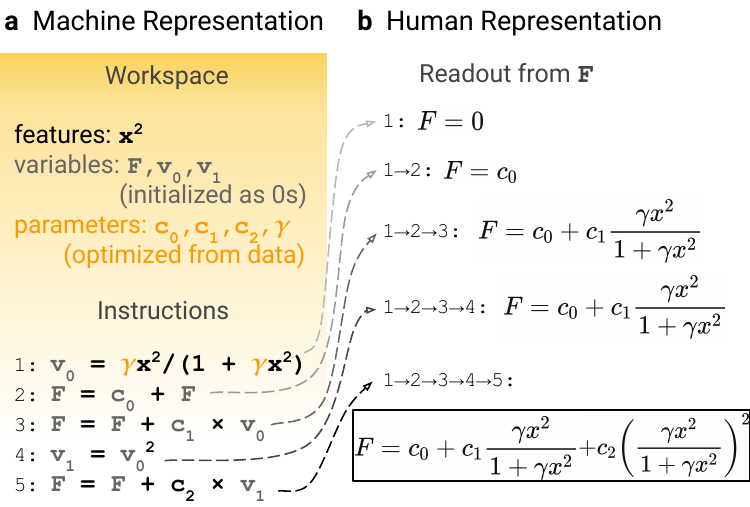}
  \caption{
  {\bf Representation of symbolic functionals.}
  {\bf a} Machine representation of the symbolic functionals. The B97 enhancement factor is represented as five consecutive  instructions operating on a workspace.
  {\bf b} The intermediate readout from $F$ during the consecutive execution of the instructions. The final readout (boxed) is the enhancement factor.
  }
  \label{workspace}
\end{figure*}

In order to represent XC functionals using instructions, we define a \textit{workspace} containing \textit{features}, \textit{variables}, and \textit{parameters}. The \textit{features} represent the input of the functional form (enhancement factors) such as electron density, density gradient, etc.
The \textit{variables} store intermediate variables during functional execution, and are all initialized to zero. $F$ is a special variable in the workspace, whose value is taken as the resulting enhancement factor after all the instructions are executed.
The \textit{parameters} represent scalars that should be optimized by fitting to the training dataset after the symbolic form is determined. Then these parameters are considered constants in the functional evaluations.
This definition of workspace mimics the memory space when using a computer program to evaluate the density functional.

We use the B97 functional as an example to illustrate this representation. In the B97 functional, enhancement factors $F_\text{x}$, $F_\text{c-ss}$ and $F_\text{c-os}$ take the form $F^\text{B97} = c_0 + c_1 u + c_2 u^2$, where the auxiliary quantity $u = \gamma x^2 / (1 + \gamma x^2)$ is a finite-domain transform of the reduced density gradient $x = 2^{1/3} |\nabla \rho| / \rho^{4/3}$. 
In Fig.~\ref{workspace}a, the B97 enhancement factor is written as 5 consecutive mathematical instructions operating on a workspace composed of 1 feature (the square of reduced density gradient, $x^2$), 4 parameters ($c_0$, $c_1$, $c_2$ and $\gamma$) and 3 variables ($v_0$, $v_1$, and $F$).
Then we execute the instructions consecutively and then read out the final value from $F$. Fig.~\ref{workspace}b lists the intermediate readout from $F$ to explain how $F$ changes during the execution. After the execution of all the instructions, the readout from the variable $F$ equals to $F^\text{B97} = c_0 + c_1 u + c_2 c^2$, where $u = \frac{\gamma x^2}{1 + \gamma x^2}$. In this paper, we consider three types of instructions: arithmetic operations (e.g. addition), power operations (e.g. square), and building blocks from existing functional forms (e.g. the $\gamma [\cdot] / (1 + \gamma [\cdot])$ operation in the B97 form), see Table \ref{instructions} for details. We remark that throughout this work we do not presume any parameters to be linear (e.g. parameters $c_0$, $c_1$ and $c_2$ in $F^\text{B97}$), and we generally treat all parameters as nonlinear parameters during the parameter optimization process.

\begin{table}[!h]
    \centering
    \begin{tabular}{p{1.5in}p{1.5in}}
    \hline
      \multicolumn{2}{l}{Arithmetic operations:} \\
      $s = p + q$ & $s = p - q$ \\
      $s = p \times q$ & $s = p / q$ \\
      \multicolumn{2}{l}{$s = s + p \times q$ $^a$} \\
    \hline
      \multicolumn{2}{l}{Power operations:} \\
      \multicolumn{2}{l}{$s = p^n, \, n \in \{2, 3, 4, 6, 1/2, 1/3 \}$} \\
    \hline
      \multicolumn{2}{l}{Building blocks from existing functionals$^b$:} \\
      \multicolumn{2}{l}{$s = \gamma p / (1 + \gamma p)$, $\gamma$ is a parameter} \\
    \hline
    \end{tabular}
    
    \footnotesize{$^a$ This instruction denotes the s += p $\times$ q instruction, where += is the addition-assignment operator in common programming languages. The inclusion of this instruction can reduce the number of instructions required to represent functionals such as B97 or $\omega$B97M-V.}
    
    \footnotesize{$^b$ This instruction comes from the B97 functional. Several other building blocks are also explored but are found to play insignificant roles, see SM for details.}
    \caption{{\bf Instructions for symbolic functional representation.} We consider arithmetic operations, powers and building blocks from existing functionals. $p$, $q$ and $s$ denote symbols in the workspace, where the right hand side symbols ($p$, $q$) can be any features, parameters or variables, and the left hand side symbol ($s$) can be any variables.  A subset of 4 instructions ($s = p + q$; $s = s + p \times q$; $s = p^2$; $s = \gamma p / (1 + \gamma p)$) is used in proof-of-principles calculations for the B97 exchange functional.}
    \label{instructions}
\end{table}

The workspace and instructions define the search space of the symbolic functionals.
Then we chose a set of four mutation rules to modify a symbolic functional: the insertion of a new instruction, the removal of an existing instruction, the change of operation in an instruction and the change of argument in an instruction. These mutation rules enable the evolutionary search procedure to generate new symbolic forms starting from existing ones, thus exploring the search space of functional forms.

\hspace{1cm}

\noindent {\it Symbolic functional evolutionary search}

After the symbolic representation of functionals is determined, the problem of searching functionals is transformed into a combinatorial optimization of instructions, features, variables and parameters. We designed a framework, SyFES, to construct the symbolic functional forms that best fit the data (Fig.~\ref{workflow}).
In particular, the regularized evolution algorithm~\cite{real2019regularized} maintains and evolves a population of \textit{individuals} based on the \textit{fitness} of individuals. In each iteration, a random subset of individuals is drawn from the entire population, and the individual with the highest fitness is chosen (tournament selection) as the parent, which is then mutated to generate a child and added back to the population. When the size of the population grows beyond a limit, the oldest individual is removed from the population. Compared to standard genetic algorithms, which usually remove the individual with the lowest fitness, the regularized evolution algorithm introduces aging to avoid individuals with good fitness but poor robustness to stay in the population for an excessive amount of time. It leads to better exploration of the search space and better generalization of the best individuals it found. 

In this work, a symbolic functional form is treated as an individual. The initial population can be the empty functional or existing functionals.
Each time a new functional is obtained, its scalar parameters are determined by minimizing the training error $J_\text{train}$.
We adopt the weighted-root-mean-squared-deviation (WRMSD) from Ref.~\citen{mardirossian2016omega} as the objective function,
\begin{equation}
    J_S = \sqrt{ \frac{1}{N} \sum_{i\in S} w_i (E_i - E^\text{ref}_i)^2 },
\label{loss}
\end{equation}
where $N$ is the total number of data points in the dataset $S$ (training, validation or test). $E_i$ and $E^\text{ref}_i$ are the $i$-th energetic data and its reference value, respectively. $E_i$ can generally be computed from the DFT total energies ($E_\text{tot}$ in equation~(\ref{etot})) of one or more molecules. For instance, a $E_i$ that corresponds to isomerization energy is computed by taking the difference of $E_\text{tot}$ between two isomers.
$w_i$ is the sample weight that accounts for the different scale of energetic data in different subsets.

After the scalar parameters are determined, SyFES records the fitness of the new functional as $-J_\text{val}$ which is used later in the tournament selection.
As the procedure advances, the population gradually evolves to functionals with higher fitness (lower validation error). The functional form with the highest fitness is selected as the final output from SyFES.

In the following two subsections we present two demonstrations of the SyFES framework. In the first proof-of-principle demonstration, SyFES is capable of finding a known functional (B97 exchange functional) from scratch. In the second demonstration, SyFES can evolve from an existing functional ($\omega$B97M-V) to a novel functional form GAS22, which has better performance (test error) on the MGCDB84 dataset. In both demonstrations, we use the electron densities evaluated from the $\omega$B97M-V functional as input to equation~(\ref{etot}) to evaluate each new functional form to accelerate the search. At the end, we present self-consistent DFT calculations using GAS22.

\hspace{1cm}

\noindent {\bf Rediscovery of B97 exchange functional}

As a proof-of-principle demonstration, we apply SyFES to a case where the ground truth functional is known. In particular, we adopt the TCE subset of MGCDB84 that contains 947 data points for thermodynamics of molecules, and we randomly partition the 947 data points into training, validation and test set which contains 568, 189 and 190 data points, respectively. We set the reference energies in equation~(\ref{loss}) to the total energies evaluated using the B97 exchange functional. In this demonstration, we verify whether SyFES is capable of finding the B97 enhancement factor $F^\text{B97}$ or its equivalence.

We design the search space to be functionals that can be represented by less or equal to six instructions from a set of four instructions $s = p + q$, $s = s + p \times q$, $s = p^2$ and $s = \gamma p / (1 + \gamma p)$. The workspace contains one feature ($x^2$), four parameters and three variables. The number of parameters is chosen corresponding to the parameters $c_0$, $c_1$, $c_2$ and $\gamma$, as shown in Fig.~\ref{workspace}a. Except for $\gamma$ which is bound to the finite-domain transform, all other parameters, together with features and variables are free to use as arguments in any instructions.

\begin{figure}[!h]
  \centering
  \includegraphics[width=3in]{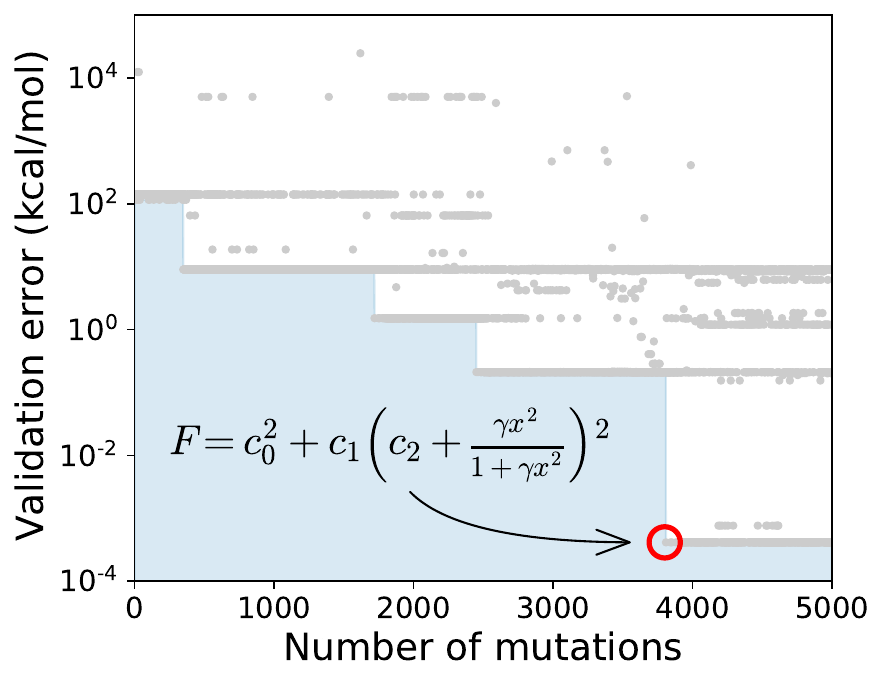}
  \caption{{\bf Validation error of symbolic functionals generated by SyFES starting from an empty functional (no instruction).} The total energies evaluated using the B97 exchange functional are used as reference energies for training and evaluation of functionals. Each grey dot represents one functional form generated by SyFES. The blue area represents the cumulative minimum validation error up to a certain number of mutations. After less than 4000 mutations, an equivalent form of B97 enhancement factor (red circle) is obtained.
  }
  \label{b97}
\end{figure}

We ran SyFES starting from scratch, where the initial functional contains no instructions and thus constantly outputs zero. Fig.~\ref{b97} shows the validation error as a function of the number of mutations, where each dot corresponds to a functional form. After less than 4000 iterations, SyFES was able to find a functional form with validation error $J_\text{val}=4.2 \times 10^{-4}$ kcal/mol and test error $J_\text{test}=3.7 \times 10^{-4}$ kcal/mol. The functional has the following symbolic form
\begin{equation}
    F = c_0^2 + c_1 \left( c_2 + \frac{\gamma x^2}{1 + \gamma x^2} \right)^2
\end{equation}
with $c_0 = 0.8504$, $c_1 = 0.7480$, $c_2 = 0.3394$, $\gamma = 0.0040$. It is easy to verify that this functional form is equivalent to the B97 one: $F_\text{x}^\text{B97} = c_0 + c_1 \frac{\gamma x^2}{1 + \gamma x^2} + c_2 \left(\frac{\gamma x^2}{1 + \gamma x^2}\right)^2$ with $c_0 = 0.8094$, $c_1 = 0.5073$, $c_2 = 0.7481$ and $\gamma = 0.004$. This study demonstrates that SyFES can find existing simple symbolic functional forms from scratch.

\hspace{1cm}

\noindent {\bf Evolving from the $\omega$B97M-V functional}

Now we turn to the main result of this work, where we demonstrate that SyFES is capable of evolving from existing functional forms to novel functional forms. We use the $\omega$B97M-V functional as a starting point. Its enhancement factors can be written as power series in two variables
\begin{equation}
    F^\text{$\omega$B97M-V} = \sum_i \sum_j c_{ij} w^i u^j,
\label{fwb97mv}
\end{equation}
where $w$ is an auxiliary quantity related to the kinetic energy density $\tau$, $u$ is the auxiliary quantity related to reduced density gradient $x$ as in the case of the B97 functional (see SM for detailed definitions). $c_{ij}$ are linear coefficients for the power series of $w$ and $u$. In the $\omega$B97M-V functional, the exchange enhancement factor $F^\text{$\omega$B97M-V}_\text{x}$ includes $c_{00}$, $c_{10}$ and $c_{01}$ terms; the same-spin correlation enhancement factor $F^\text{$\omega$B97M-V}_\text{c-ss}$ includes $c_{00}$, $c_{10}$, $c_{20}$, $c_{43}$, and $c_{04}$ terms; the opposite-spin correlation enhancement factor $F^\text{$\omega$B97M-V}_\text{c-os}$ includes $c_{00}$, $c_{10}$, $c_{20}$, $c_{60}$, $c_{21}$ and $c_{61}$ terms. These terms were determined through a best subset selection procedure using the training and validation set of MGCDB84~\cite{mardirossian2016omega}. Overall, there are 15 linear parameters ($c_{ij}$'s) and 3 nonlinear parameters (the $\gamma$ parameters in the definition of $u$ for $F_\text{x}$, $F_\text{c-ss}$ and $F_\text{c-os}$, respectively) in the semilocal part of the $\omega$B97M-V functional. The linear parameters are determined by performing linear regression on the MGCDB84 training set, and the nonlinear parameters are directly taken from previous studies.

To use the $\omega$B97M-V functional as the starting point in SyFES, we first represent it as consecutive instructions. We adopted all instructions shown in Table~\ref{instructions}, which leads to a significantly larger search space than the proof-of-principles demonstration on the B97 exchange functional. In addition to instructions used for searching the B97 exchange functional, we also include subtraction, division, and powers with additional exponents. Besides arithmetic operations, we also explored using existing functional building blocks from existing functionals, but we found that those forms are too restrictive to be selected by the SyFES. Using such a choice of instruction set, $F^\text{$\omega$B97M-V}_\text{x}$, $F^\text{$\omega$B97M-V}_\text{c-ss}$, and $F^\text{$\omega$B97M-V}_\text{c-os}$ can be represented using 6, 11 and 11 instructions, respectively (see SM for the symbolic representation of $\omega$B97M-V).

\begin{figure}[!h]
  \centering
  \includegraphics[width=5in]{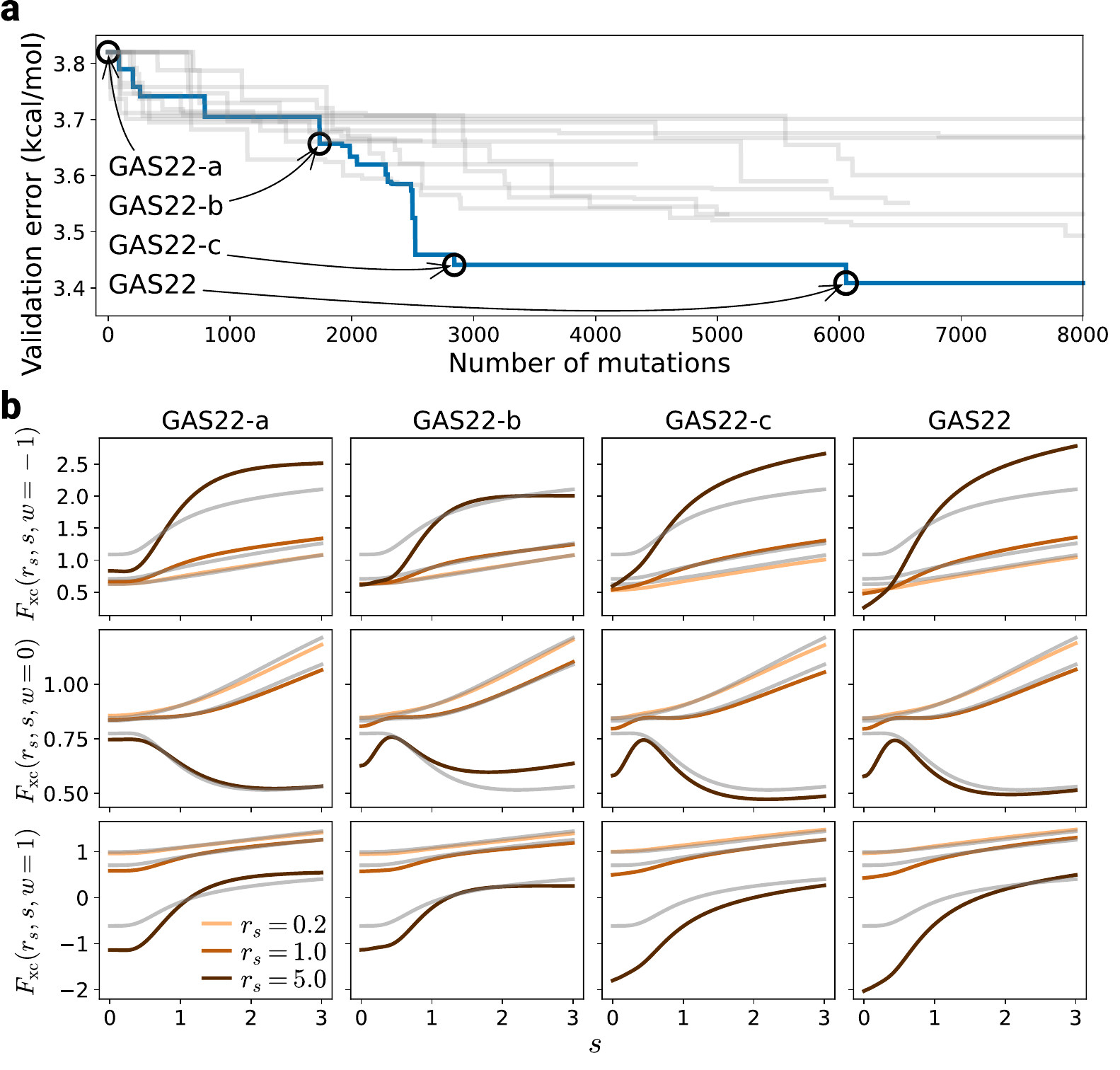}
  \caption{
  {\bf Evolving functional forms starting from $\omega$B97M-V.} {\bf a} Validation error of symbolic functionals generated by SyFES. Different curves denotes different independent evolutions, the evolution trajectory with the optimal validation error individual is colored blue. {\bf b} XC enhancement factors $F_\text{xc}$ (see text) of 4 snapshots of functionals in the best evolution. $F_\text{xc}$ are plotted as functions of dimensionless quantities $s = x / 2 (3\pi^2)^{1/3}$ and $w$ at different Wigner-Seitz radius $r_s = (3 / 4 \pi \rho) ^{1/3}$. Grey curves denote the enhancement factors of $\omega$B97M-V functional.}
  \label{evolution}
\end{figure}

Based on this representation, we performed a set of 12 independent evolutions starting from the $\omega$B97M-V functional. These evolutions explored 29628 functional forms in total. Each curve in Fig.~\ref{evolution}a presents the cumulative minimum validation error of symbolic functionals explored in one evolution. Once a new functional form is obtained with lower validation error than all the previous functionals, the plot has a step down. The evolution leading to the functional with lowest validation error is marked in blue. The figure shows that SyFES is able to generate new functional forms with decreasing validation error as the evolution proceeds. For comparison, we also performed a set of random search studies that randomly mutate functionals from the population without the tournament selection. The random search cannot effectively obtain functionals with improved performance (more details in SM).

Now we turn to the behavior of functionals. We denote the best functional obtained by the end of the evolution as GAS22. And we pick three precedent functional forms GAS22-a, GAS22-b and GAS22-c along the best trajectory in Fig.~\ref{evolution}a. In order to illustrate their numerical behavior, in Fig.~\ref{evolution}b we present their XC enhancement factor $ F_\text{xc} = e_\text{xc} / e^\text{LDA}_\text{xc}  = ( e^\text{LDA}_\text{x} F_\text{x} + e^\text{LDA}_\text{c-ss} F_\text{c-ss} + e^\text{LDA}_\text{c-os} F_\text{c-os} )/( e^\text{LDA}_\text{x} + e^\text{LDA}_\text{c-ss} + e^\text{LDA}_\text{c-os} ) $, which characterize the deviation of XC energy density from that of the LDA. Since we are training a meta-GGA functional, $F_\text{xc}$ would depend on density, density gradient and kinetic energy density. The $F_\text{xc}$ of $\omega$B97M-V are plotted in grey for comparison. The first, second, and third row show the behavior of $F_\text{xc}$ at $w = \{-1, 0, 1\}$, which correspond to chemical bonds with weak, metallic and covalent characters, respectively~\cite{sun2013density}. In each individual subplot, we plot $F_\text{xc}$ as a function of $s = x / 2 (3\pi^2)^{1/3}$ at multiple values of Wigner-Seitz radius $r_s$. $s$ is proportional to the reduced density gradient $x$ and is a common auxiliary quantity used in literature for analyzing density functionals. Normal physical systems usually have $s$ between 0 and 3~\cite{perdew1996generalized}. The Wigner-Seitz radius $r_s = (3 / 4 \pi \rho) ^{1/3}$ characterizes the electron density, where a larger value of $r_s$ corresponds to lower electron density. Based on the plots, one can see that the GAS22 (brown curves) differ from $\omega$B97M-V (grey curves) in a few regions. The first regime involves small density gradient ($s < 1$), where $F_\text{xc}$ of GAS22 tends to be lower than $\omega$B97M-V. The second regime involves weak bonds ($w = -1$) and small electron density ($r_s = 5$), where $F_\text{xc}$ of GAS22 tends to be higher than $\omega$B97M-V.

After simplification, the final symbolic form of GAS22 is,
\begin{align}
    F_\text{x} &= 0.862 + 0.937 u + 0.318 w \label{fx} \\
    F_\text{c-ss} &=  u - 4.108 w  - 5.242 w^{2} - 1.766 u^6 + 7.538 w^{4} u^6 \label{fc-ss} \\
    F_\text{c-os} &= 0.805 + 7.989 w^{2} -7.548 w^{6} + 2.001 w^{6} \sqrt[3]{x^{2}} - 1.761 w^{2} \sqrt[3]{x^{2}} \label{fcos}
\end{align}
where $u=\gamma x^2 / (1 + \gamma x^2)$, similar to the B97 and the $\omega$B97M-V functional, with $\gamma = 0.00384$ in $F_\text{x}$ and $\gamma = 0.469$ in $F_\text{c-ss}$. The exchange enhancement factor in equation~(\ref{fx}) is symbolically identical and numerically similar (Fig. S1 last column) to the $\omega$B97M-V functional, which indicates that the $\omega$B97M-V exchange enhancement factor $F^{\omega{\text{B97M-V}}}_\text{x}$ may be accurate enough for depicting the exchange. In fact, the best subset selection presented in Ref.~\citen{mardirossian2016omega} only selected three lowest-order terms for the definition of $F^{\omega\text{B97M-V}}_\text{x}$, indicating that the exchange functional is easily captured by the form of 2D power series in $u$ and $w$. SyFES recognized this and maintained the symbolic form of the exchange functional. The same-spin correlation enhancement factor in equation~(\ref{fc-ss}) still assumes the form of power series in two variables, but the orders are no longer those in $\omega$B97M-V, indicating that the symbolic regression is capable of applying minor symbolic modifications to existing forms for lower error. The most striking difference is found in the opposite-spin correlation enhancement factor in equation~(\ref{fcos}), which contains a novel $x^{3/2}$ term that is completely outside of the space spanned by power series in $u$ and $w$ as in equation~(\ref{fwb97mv}). It highlights the power of SyFES in discovery novel functional forms from data. To assess the performance of the functional, we apply it to the test set of MGCDB84, which was not used during the training and validation of functional forms. The test error of GAS22 is 3.585 kcal/mol, a 15\% percent improvement over the $\omega$B97M-V functional (4.237 kcal/mol).

\hspace{1cm}

\noindent {\bf Self-consistent calculations using GAS22}

\begin{figure}[!h]
  \centering
  \includegraphics[width=7in]{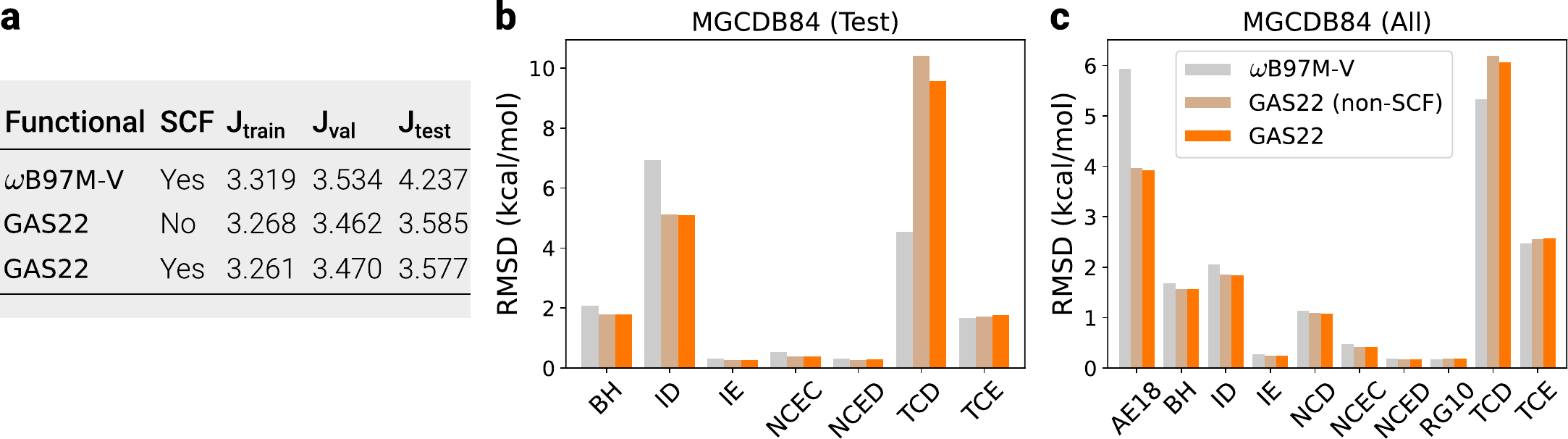}
  \caption{
  {\bf Performance of the functionals on the MGCDB84 dataset.}
  {\bf a} Training, validation and test WRMSD (kcal/mol) of GAS22 functional on the MGCDB84 dataset. Non-SCF results are evaluated using $\omega$B97M-V density.
  {\bf b} RMSD on the test set of MGCDB84 dataset. {\bf c} RMSD on the entire MGCDB84 dataset. AE18 stands for atomic energy, BH stands for barrier height, ID stands for isomerization energies (difficult), IE stands for isomerization energies (easy),  NCEC stands for non-covalent clusters (easy), NCED stands for non-covalent dimers (easy), NCD stands for non-covalent dimers (difficult), RG10 stands for rare-gas dimers, TCD stands for thermochemistry (difficult), TCE stands for thermochemistry (easy).}
  \label{rmsd}
\end{figure}

So far all the results presented are based on non-self-consistent calculations on $\omega$B97M-V densities. In order to evaluate the performance of GAS22 in realistic DFT calculations, we performed self-consistent-field (SCF) calculations where the functional derivatives are computed using automatic differentiation. Fig.~\ref{rmsd}a presents the training, validation and test errors of GAS22 after performing SCF calculations. SCF results are very similar to non-SCF ones, demonstrating good numerical stability of GAS22 found by SyFES.

The training, validation and test error were computed as WRMSD defined in equation~\ref{loss}, with weights reported in Ref.~\citen{mardirossian2016omega} (see Methods for additional details). Since the same weights were used to compute the training and validation errors for the development of $\omega$B97M-V functional, we use WRMSD on the test set as a general summary of the overall performance. We understand the insufficiency of using a scalar to measure the performance of functionals on diverse subsets. Therefore, to further benchmark the performance of GAS22 on different types of molecules, in Fig.~\ref{rmsd}b and c we report the root mean square deviation (RMSD) of the GAS22 and the $\omega$B97M-V functional on subsets of MGCDB84.
We see that GAS22 outperforms $\omega$B97M-V for most subsets. The only subset where GAS22 shows a less favorable RMSD than $\omega$B97M-V is the TCD subset, which is composed of strongly-correlated molecules. The comparison between the SCF results of GAS22 and $\omega$B97M-V demonstrates that despite not using SCF calculations during the evolution, SyFES is capable of finding functional forms with good performance in realistic SCF calculations.

\section*{Discussion}

We proposed a novel ML approach for developing accurate XC functionals, in contrast to conventional human-designed symbolic functionals and other ML generated numerical functionals. SyFES can automatically search functionals that best fit the given dataset from a large, nonlinear, symbolic functional space. As demonstrated, it is capable of finding simple existing functionals from scratch, as well as evolving an existing functional to a better performing functional. Despite the fact that the search procedure is conducted by computers, it is worth noting that the form of functionals produced by SyFES have similar simplicity as functionals designed by humans in the past few decades -- symbolic forms with a manageable amount of parameters. Thus they are amenable to all the interpretation methods scientists developed to examine the properties of functionals and understand their 
limitations. Meanwhile, the computational cost to use and the work to implement these functionals in popular libraries like Libxc~\cite{libxc} are also the same as the conventional functionals on the same level of Jacob's ladder~\cite{perdew2001jacob}. SyFES is ML applied to developing human-readable scientific expressions and not just a blackbox prediction.

Given the ubiquity of DFT in quantum simulations, we expect many applications along this direction in chemistry, physics, and materials sciences.
This work focused on main-group chemistry using the data from the MGCDB84 dataset. But the framework itself is general and can be applied to more systems via incorporating new dataset, objective function and search space.
The design of SyFES is highly scalable on computer clusters and cloud platforms, where the mutation, training and validation of new functionals are distributed into different workers asynchronously. In this work we used up to 50 workers, but it can easily be scaled up to leverage more computing power to deal with a larger search space for more demanding problems.
We conclude with a list of promising future workstreams:
1. {\it Better dataset}. It has been proved many times that high quality and comprehensive datasets are critical to development and benchmarking of ML algorithms for target applications. For example, ImageNet~\cite{deng2009imagenet} spawned the revolution of ML in computer vision. This area is under-invested for the ML DFT community.
2. {\it Refined methodology}. Starting from the results in this paper, SyFES can be further improved in multiple dimensions, e.g. search space design, feature engineering, regularization via numerical techniques, symbolic constraints or adding density information.
3. {\it Richer applications in density functionals}. It includes developing highly accurate empirical XC functionals for a family of systems, pursuing the universal functional with the attempts to include more exact conditions, constructing accurate kinetic energy functionals for large scale systems in orbital-free DFT, and formulating classical DFT~\cite{evans_2016,lin2020analytical} for fluids or even crowds~\cite{mendez2018density}.

Could SyFES or it successors replace human in the development of functionals? Conversely, humans have a long history of using computers to assist scientific discovery and recent advances in ML guided mathematicians discovering new insights in topology and representation theory~\cite{davies2021advancing}. 
SyFES may help scientists focus more on the physics insights and quality of functionals. This research direction paves the way of efficient development of density functionals using advanced algorithms and computing power.

\section*{Materials and Methods}

\noindent \textbf{Dataset and objective function}

In this work we use the Main Group Chemistry Database (MGCDB84) to train and evaluate functional forms. The training, validation and test set used in this work corresponding to the training, primary test and secondary test set in Ref.~\citen{mardirossian2016omega}. The objective function defined in equation~(\ref{loss}) is identical to Ref.~\citen{mardirossian2016omega}, thus facilitating a direct comparison between the symbolic functionals obtained in this work and $\omega$B97M-V.

The MGCDB84 dataset contains 9 types of molecular energetics data: AE18 (atomic energies), NCED (non-covalent dimers (easy)), NCEC (non-covalent clusters (easy)), NCD (non-covalent dimers (difficult)), IE (isomerization energy (easy)), ID (isomerization energy (difficult)), RG10 (rare gas potential energy curves), TCE (thermochemistry (easy)), TCD (thermochemistry (difficult)), BH (barrier heights), where the difficult/easy in the names represents the presence/lack of interactions that are difficult to capture by KS-DFT, such as strong electron correlation. There are in total 4986 data points in the dataset, with accuracy estimated to be at least 10 times more accurate than best available DFT calculations. Most data points in the dataset are relative energy differences between different molecular species, and 5931 single-point DFT calculations are required to evaluate the 4986 data points.

The proof-of-principle calculations targeting B97 exchange functional uses the TCE subset (947 data points) and a standard 60\%-20\%-20\% splitting for constructing training, validation and test sets.

For evolution studies starting from $\omega$B97M-V, we used the same training, validation and test sets as Ref.~\citen{mardirossian2016omega, mardirossian2017thirty}, which contains 870, 2960 and 1150 data points, respectively. We note that this grouping is different from standard machine learning practice, which assumes the training, validation and test sets are drawn from the same distribution. In this partition, the training, validation and test set each includes different subsets of molecular properties, which poses stronger criteria for the transferability of functionals trained on the dataset. For the calculation of training, validation and test error as defined in equation \ref{loss}, we used the following weights~\cite{mardirossian2016omega} for different data types: 0.1 for TCD; 1 for TCE and AE18; 10 for NCD, ID, and BH; 100 for NCE and IE; 10,000 for RG10.

\hspace{1cm}

\noindent \textbf{Regularized evolution}

Regularized evolution are performed with a massively-parallel implementation (see SM for the software design). The implementation takes advantage of just-in-time (JIT) compilation~\cite{bradbury2018jax} to enable training of functional forms on graphics processing units (GPU).
The population (Fig.~\ref{workflow}a) of functional forms is managed in a CPU server. Each evolution utilizes 50 parallel GPU workers to evolve functionals (Fig.~\ref{workflow}b-f) asynchronously.
All evolutions adopt a maximum population size of 100 and a tournament selection size of 25, except for the evolution that targets the B97 exchange functional, which uses a tournament size of 10. 

In the symbolic representation used in this work, one functional form may have multiple equivalent symbolic representations. For example, a mutation may introduce instructions that have no effect in the current functional, although such instructions may become useful in subsequent mutations. We designed a functional equivalence checking mechanism to circumvent the duplicated training and validation of equivalent functional forms (see SM for details), which accelerate the functional search by an order of magnitude.

\hspace{1cm}

\noindent \textbf{Parameter optimization}

After mutation, the scalar parameters in the new functional form are optimized to minimize the training error $J_\text{train}$. The optimization is performed using the covariance matrix adaptation evolution strategy (CMA-ES) method ~\cite{hansen2019pycma}. A CMA-ES optimization proceeds by iteratively optimizing a population of parameters, with the covariance matrix characterizing the spread of the population updated on-the-fly during the optimization process. In general, for a given functional form, there may be different choice of parameters that all lead to low training error (i.e. multiple local minima). For each functional form, the CMA-ES optimization is performed multiple times with initial guess for parameters randomly drawn from a unit Gaussian distribution, and the parameters leading to the lowest training error are adopted. The optimizations for each functional is repeated 10 times in the evolution for B97 exchange functional and 5 times for evolutions starting from $\omega$B97M-V functional. Due to the flexible nature of functional forms generated in this work, we constrained the value of all the parameters to [-10, 10] to avoid overfitting and numerical instability.

\hspace{1cm}

\noindent \textbf{Self-consistent DFT calculations}

Self-consistent DFT calculations are performed by solving the KS equations,
\begin{equation}
    -\nabla^2/2 + v_\text{ext} + v_{H} + v_\text{xc} \psi_i(\bm{r}) = \varepsilon_i \psi_i(\bm{r}),
\label{ks_equation}
\end{equation}
where $\psi_i$ denotes the KS orbitals. $v_\text{ext}$, $v_{H}$ and $v_\text{xc}$ are external, Hartree and XC potential, respectively. The XC potential $v_\text{xc}$ is defined as the functional derivative of the XC functional $E_\text{xc}$ with respect to the electron density $\rho(\bm{r})$,
\begin{equation}
    v_\text{xc}(\bm{r}) = \frac{\delta E_\text{xc}[\rho(\bm{r})]}{\delta \rho(\bm{r})},
\label{vxc}
\end{equation}
where the density $\rho(\bm{r})=\sum_i^\text{occ} |\psi_i(\bm{r})|^2$ sums over occupied KS orbitals. For self-consistent calculations with the GAS22 functional, the functional derivative in equation~(\ref{vxc}) is evaluated through automatic differentiation using the JAX package~\cite{bradbury2018jax}.

All self-consistent DFT calculations are performed with the PySCF package~\cite{Sun2017} with a large basis set def2-QZVPPD~\cite{rappoport2010property}. The default integration grid in PySCF is adopted for the evaluation of semilocal XC energies; SG-1 prune~\cite{gill1993standard} is used for evaluating VV10~\cite{vydrov2010nonlocal} nonlocal correlation energies.

Out of the 5931 single-point DFT calculations needed to evaluate the entire MGCDB84 dataset, there are 7/8 single-point SCF calculations with the $\omega$B97M-V/GAS22 functional that did not achieve convergence, which affects the evaluation the 6/7 data points out of all the 4986 data points.
Using reference values for $\omega$B97M-V results reported in Ref.~\citen{mardirossian2016omega}, we estimate that excluding the these data points lead to less than 0.01\% change in the calculation of training, validation and test errors.

\section*{Acknowledgments}

The authors thank Kieron Burke, Narbe Mardirossian, Ekin Dogus Cubuk, Chen Liang, Hanxiao Liu and Stephan Hoyer for helpful discussion on DFT, AutoML and efficient scientific computing using JAX.

\hspace{1cm}

\noindent {\bf Author contributions:}
LL and PR conceived the original idea to search symbolic density functionals. LL concretized the project formulation and implemented the distributed infrastructure for SyFES. HM designed and implemented  SyFES and corresponding changes in PySCF. LL and AN reviewed the code. HM ran all the experiments. All authors contributed to the design of the experiments and discussion of the results. HM and LL wrote the paper.

\hspace{1cm}

\noindent {\bf Competing interests:}
Authors declare that they have no competing interests.

\hspace{1cm}

\noindent {\bf Data and materials availability:}
The code of SyFES is publicly available at \url{https://github.com/google-research/google-research/tree/master/symbolic_functionals}. And we provide an example Colab notebook to demonstrate self-consistent DFT calculations with GAS22 functional at \url{https://colab.research.google.com/github/google-research/google-research/blob/master/symbolic_functionals/colab/run_GAS22.ipynb}.

\section*{Supplementary Materials}

\section{Functional forms}

In this section we present the functional forms in main text Eq. 4-5 for general systems (which may contains spin polarization). 

The semilocal part of the exchange-correlation functional assumes the following form
\begin{equation}
    E_\text{xc}^\text{sl} = \int \left( \sum_\sigma e^\text{LDA}_{\text{x-sr}, \sigma} F_{\text{x}, \sigma} + \sum_\sigma e^\text{LDA}_{\text{c-ss}, \sigma} F_{\text{c-ss}, \sigma} + e^\text{LDA}_{\text{c-os}} F_{\text{c-os}} \right) \text{d}\bm{r}
\label{exc_f}
\end{equation}
where $\sigma \in \{\alpha, \beta\}$ is the spin index; $e^\text{LDA}_{\text{x-sr}, \sigma}$, $e^\text{LDA}_{\text{c-ss}, \sigma}$, and $e^\text{LDA}_{\text{c-os}}$ are short-range exchange, same-spin correlation and opposite-spin correlation energy densities within local (spin) density approximation. The partition of correlation energy into same-spin and opposite-spin contributions adopts the widely-used scheme proposed by Stoll et al.~\cite{stoll1978theor}. The short-range LDA exchange energy density $e^\text{LDA}_{\text{x-sr}, \sigma}$ is obtained by multiplying the LDA exchange energy density $e^\text{LDA}_{\text{x}, \sigma}$ with an attenuation function
\begin{equation}
    e^\text{LDA}_{\text{x-sr}, \sigma} = e^\text{LDA}_{\text{x}, \sigma} \left[1 - \frac{2}{3} a_\sigma \left( 2 \sqrt{\pi} \text{erf}\left(\frac{1}{a_\sigma}\right) - 3 a_\sigma + a_\sigma^3 + (2a_\sigma - a_\sigma^3) \text{exp}\left(-\frac{1}{a_\sigma^2}\right) \right) \right]
\label{ex_lda_sr}
\end{equation}
where $a_\sigma = \omega / k_{\text{F}\sigma}$ with $k_{\text{F}\sigma} = (3\pi^2 \rho)^{1/3}$ being the Fermi wave vector and $\omega$ being the range-separation parameter.

$F_{\text{x}, \sigma}$, $F_{\text{c-ss}, \sigma}$ and $F_{\text{c-os}}$ are the exchange, same-spin correlation and opposite-spin correlation enhancement factors that depends on reduced density gradient and kinetic energy density
\begin{equation}
    F_{\text{x}, \sigma} = F_{\text{x}, \sigma}(x^2_\sigma, w_\sigma), \, F_{\text{c-ss}, \sigma} = F_{\text{c-ss}, \sigma}(x^2_\sigma, w_\sigma), \, F_\text{c-os} = F_\text{c-os}(x^2_\text{ave}, w_\text{ave})
\end{equation}
where $x_\sigma = \frac{|\grad \rho_\sigma|}{\rho_\sigma^{4/3}}$ denotes the reduced density gradient. $w_\sigma$ is an auxiliary quantity that depends on kinetic energy density $\tau_\sigma = \frac{1}{2} \sum_i^{\text{occ}} |\grad \psi_{i\sigma}|^2$, with $\psi$'s being Kohn-Sham orbitals and the summation runs over occupied Kohn-Sham orbitals. In particular, $w_\sigma = (t_\sigma - 1) / (t_\sigma + 1)$, with $t_\sigma = \tau^\text{HEG}_\sigma / \tau_\sigma$ where $\tau^\text{HEG}_\sigma = \frac{3}{10} (6 \pi^2)^{2/3} \rho_\sigma^{5/3}$ is the kinetic energy density of homogenous electron gas (HEG). The opposite-spin correlation enhancement factor $F_{\text{c-ss}, \sigma}$ depends on spin-averaged version of $x^2$ and $w$, defined as $x^2_\text{ave} = \frac{1}{2}(x^2_\alpha + x^2_\beta)$ and $w_\text{ave} = (t_\text{ave} - 1) / (t_\text{ave} + 1)$ with $t_\text{ave} = \frac{1}{2} (t_\alpha + t_\beta)$. We note that the form of input features for enhancement factors defined here are widely-used in B97-inspired functional forms.

The nonlocal part of the exchange-correlation functional contains the short-range exact-exchange $E_\text{x-sr}^\text{exact}$, long-range exact exchange $E_\text{x-lr}^\text{exact}$ and VV10 nonlocal correlation $E_\text{c}^\text{VV10}$. The short-range and long-range exact exchange assume the following form
\begin{equation}
    E_\text{x-sr}^\text{exact}[\rho] = -\frac{c_\text{x}}{2} \sum_\sigma \sum_{i,j}^\text{occ} \int \int \psi^*_{i\sigma}(\bm{r}_1) \psi^*_{j\sigma}(\bm{r}_2) \frac{\text{erfc}(\omega r)}{r} \psi_{j\sigma}(\bm{r}_1) \psi_{i\sigma}(\bm{r}_2) d\bm{r}_1 d\bm{r}_2
\end{equation}
\begin{equation}
    E_\text{x-lr}^\text{exact}[\rho] = -\frac{1}{2} \sum_\sigma \sum_{i,j}^\text{occ} \int \int \psi^*_{i\sigma}(\bm{r}_1) \psi^*_{j\sigma}(\bm{r}_2) \frac{\text{erf}(\omega r)}{r} \psi_{j\sigma}(\bm{r}_1) \psi_{i\sigma}(\bm{r}_2) d\bm{r}_1 d\bm{r}_2
\end{equation}
where $r = |\bm{r}_1 - \bm{r}_2|$ and $\omega$ is a range-separation parameter controlling the characteristic length scale for range separation. Note that there is a prefactor $c_\text{x}$ controlling the amount of short-range exact exchange used in the functional form. The exchange functional used in this work thus behaves as purely exact exchange in long range and a mixture of semilocal and exact exchange in short range.  The VV10 nonlocal correlation $E_\text{c}^\text{VV10}$ assumes the form
\begin{equation}
    E_\text{c}^\text{VV10}[\rho] = \int \rho(\bm{r}_1) \left[ \frac{1}{32} \left[ \frac{3}{b^2} \right]^{3/4} + \frac{1}{2} \int \rho(\bm{r}_2) \Phi(\bm{r}_1, \bm{r}_2; {b, C}) d\bm{r}_2 \right] d\bm{r}_1
\end{equation}
where integration kernel $\Phi$ depends on two empirical parameters $b$ and $C$ (see Ref. \cite{vydrov2010nonlocal} for expression). We keep all the empirical parameters in nonlocal terms to be identical to those in $\omega$B97M-V, namely $\omega = 0.3$, $c_\text{x} = 0.15$, 
$b = 6$ and $C = 0.01$.

\pagebreak

\section{Evolution of symbolic functional forms}

The simplified mathematical forms of functional forms shown in Fig. 3 of the main text is shown below. $c$'s and $\gamma$ are parameters. The same symbol (e.g. $c_0$) in different enhancement factors of the same functional represent different parameters. See Table S1 for numerical values for the parameters in the GAS22 functional.

GAS22-a ($\omega$B97M-V):
$$F_\text{x} = c_{0} + c_{1} w + \frac{c_{2} \gamma x^{2}}{1 + \gamma x^{2}}$$
$$F_\text{c-ss} = c_{0} + c_{1} w + c_{2} w^{2} + \frac{c_{3} \gamma^{4} x^{8}}{\left(1 + \gamma x^{2}\right)^{4}}  + \frac{c_{4} \gamma^{3} w^{4} x^{6}}{\left(1 + \gamma x^{2}\right)^{3}}$$
$$F_\text{c-os} = c_{0} + c_{1} w + c_{2} w^{2}  + c_{3} w^{6} + \frac{c_{4} \gamma w^{2} x^{2}}{1 + \gamma x^{2}} + \frac{c_{5} \gamma w^{6} x^{2}}{1 + \gamma x^{2}}$$

GAS22-b:
$$F_\text{x} = c_{0} + c_{1} w + \frac{c_{2} \gamma x^{2}}{1 + \gamma x^{2}}$$
$$F_\text{c-ss} = c_{1} w + c_{2} w^{2} + \frac{c_{3} \gamma^{6} w^{4} x^{12}}{\left(1 + \gamma x^{2}\right)^{6}} + \frac{c_{4} \gamma^{4} x^{8}}{\left(1 + \gamma x^{2}\right)^{4}} + \frac{\gamma x^{2}}{1 + \gamma x^{2}}$$
$$F_\text{c-os} = c_{0} + c_{2} w^{2} + c_{3} w^{6} + \frac{c_{4} \gamma w^{6} x^{2}}{1 + \gamma x^{2}} + \frac{c_{5} \gamma w^{2} x^{2}}{1 + \gamma x^{2}}$$

GAS22-c:
$$F_\text{x} = c_{0} + c_{1} w + \frac{c_{2} \gamma x^{2}}{1 + \gamma x^{2}}$$
$$F_\text{c-ss} = c_{1} w + c_{2} w^{2} + \frac{c_{3} \gamma^{6} w^{4} x^{12}}{\left(1 + \gamma x^{2}\right)^{6}} + \frac{c_{4} \gamma^{6} x^{12}}{\left(1 + \gamma x^{2}\right)^{6}} + \frac{\gamma x^{2}}{1 + \gamma x^{2}}$$
$$F_\text{c-os} = c_{0} + c_{2} w^{2} + c_{3} w^{6} + c_{4} w^{6} \sqrt[3]{x^{2}} + c_{5} w^{2} \sqrt[3]{x^{2}}$$

GAS22:
$$F_\text{x} = c_{0} + c_{1} w + \frac{c_{2} \gamma x^{2}}{1 + \gamma x^{2}}$$
$$F_\text{c-ss} = c_{1} w + c_{2} w^{2} + \frac{c_{3} \gamma^{6} w^{4} x^{12}}{\left(1 + \gamma x^{2}\right)^{6}} + \frac{c_{4} \gamma^{6} x^{12}}{\left(1 + \gamma x^{2}\right)^{6}} + \frac{\gamma x^{2}}{1 + \gamma x^{2}}$$
$$F_\text{c-os} = c_{0} + c_{2} w^{2} + c_{3} w^{6} + c_{4} w^{6} \sqrt[3]{x^{2}} + c_{5} w^{2} \sqrt[3]{x^{2}}$$

\begin{table}[!h]
    \centering
    \begin{tabular}{l l}
    \hline
      \multicolumn{2}{c}{$F_\text{x}$} \\
        $c_0$ & $0.862139736374172$  \\
        $c_1$ & $0.317533683085033$ \\
        $c_2$ & $0.936993691972698$ \\
        $\gamma$ & $0.003840616724010807$ \\
    \hline
      \multicolumn{2}{c}{$F_\text{c-ss}$} \\
        $c_1$ & $-4.10753796482853$ \\
        $c_2$ & $-5.24218990333846$ \\
        $c_3$ & $7.5380689617542$ \\
        $c_4$ & $-1.76643208454076$ \\
        $\gamma$ & $0.46914023462026644$ \\
    \hline
      \multicolumn{2}{c}{$F_\text{c-os}$} \\
        $c_0$ & $0.805124374375355$ \\
        $c_2$ & $7.98909430970845$ \\
        $c_3$ & $-7.54815900595292$ \\
        $c_4$ & $2.00093961824784$ \\
        $c_5$ & $-1.76098915061634$ \\
    \hline
    \end{tabular}
    
    \caption{Parameters in the GAS22 functional.}
\end{table}

\pagebreak

\section{Symbolic representations of density functionals}

As stated in the Table 1 of the main text, the instructions used in this work include 3 categories: arithmetic operations, power operations and building blocks from existing functionals. For the category of building blocks of existing functionals, we considered a few additional instructions in addition to the $\gamma x / (1 + \gamma x)$ presented in Table 1, including PBE exchange enhancement factor $F_\text{x}^\text{PBE}$~\cite{perdew1996generalized}, RPBE exchange enhancement factor $F_\text{x}^\text{RPBE}$~\cite{hammer1999improved}, B88 exchange enhancement factor $F_\text{x}^\text{B88}$~\cite{becke1988density} and PBE correlation energy functional $E_\text{c}^\text{PBE}$~\cite{perdew1996generalized}. 

We design the probability such that similar instructions receive identical probabilities and probabilities distribute evenly among different types of instructions. For the 5 arithmetic operations, each operation receive a probability of 0.06; for the 6 power instructions, each receive a probability of 0.05; u transform receive a probability of 0.1, and the other 4 building block receive a 0.075 each. 

Symbolic representation of $\omega$B97M-V:

%
\begin{algorithm}
\caption{$F^{\omega\text{B97M-V}}_\text{x}$}
\SetKwInput{KwFeature}{Features}
\SetKwInput{KwVariable}{Variables}
\SetKwInput{KwParameter}{Parameters}
\DontPrintSemicolon

  \KwFeature{$w$, $x^2$}
  \KwVariable{$F$, $v_0$, $v_1$}
  \KwParameter{$\gamma$, $c_{00}$, $c_{10}$, $c_{01}$}

  \SetKwProg{Fn}{Instructions}{:}{}
  \Fn{}{
        $v_0 = \gamma x^2 / (1 + \gamma x^2)$\;
        $F = c_{00} + F$\;
        $v_1 = c_{10} \times w$\;
        $F = F + v_1$\;
        $v_1 = c_{01} \times v_0$\;
        $F = F + v_1$\;
        \KwRet $F$\;
  }
\end{algorithm}

%
\begin{algorithm}
\caption{$F^{\omega\text{B97M-V}}_\text{c-ss}$}
\SetKwInput{KwFeature}{Features}
\SetKwInput{KwVariable}{Variables}
\SetKwInput{KwParameter}{Parameters}
\DontPrintSemicolon

  \KwFeature{$w$, $x^2$}
  \KwVariable{$F$, $v_0$, $v_1$, $v_2$, $v_3$}
  \KwParameter{$\gamma$, $c_{00}$, $c_{10}$, $c_{20}$, $c_{43}$, $c_{04}$}

  \SetKwProg{Fn}{Instructions}{:}{}
  \Fn{}{
        $v_0 = \gamma x^2 / (1 + \gamma x^2)$\;
        $F = c_{00} + F$\;
        $F += c_{10} \times w$\;
        $v_1 = w^2$\;
        $F += c_{20} \times v_1$\;
        $v_1 = w^4$\;
        $v_2 = v_0^3$\;
        $v_3 = v_3 \times v_2$\;
        $F += c_{43} \times v_3$\;
        $v_2 = v_0^4$\;
        $F += c_{04} \times v_2$\;
        \KwRet $F$\;
  }
\end{algorithm}

%
\begin{algorithm}
\caption{$F^{\omega\text{B97M-V}}_\text{c-os}$}
\SetKwInput{KwFeature}{Features}
\SetKwInput{KwVariable}{Variables}
\SetKwInput{KwParameter}{Parameters}
\DontPrintSemicolon

  \KwFeature{$w$, $x^2$}
  \KwVariable{$F$, $v_0$, $v_1$, $v_2$, $v_3$}
  \KwParameter{$\gamma$, $c_{00}$, $c_{10}$, $c_{20}$, $c_{21}$, $c_{60}$, $c_{61}$}

  \SetKwProg{Fn}{Instructions}{:}{}
  \Fn{}{
        $v_0 = \gamma x^2 / (1 + \gamma x^2)$\;
        $F = c_{00} + F$\;
        $F += c_{10} \times w$\;
        $v_1 = w^2$\;
        $F += c_{20} \times v_1$\;
        $v_3 = v_1 \times v_0$\;
        $F += c_{21} \times v_1$\;
        $v_1 = w^6$\;
        $F += c_{60} \times v_1$\;
        $v_3 = v_1 \times v_0$\;
        $F += c_{61} \times v_3$\;
        \KwRet $F$\;
  }
\end{algorithm}

\pagebreak

\section{Enhancement factors of symbolic functionals}

\begin{figure*}[!h]
  \centering
  \includegraphics[width=5in]{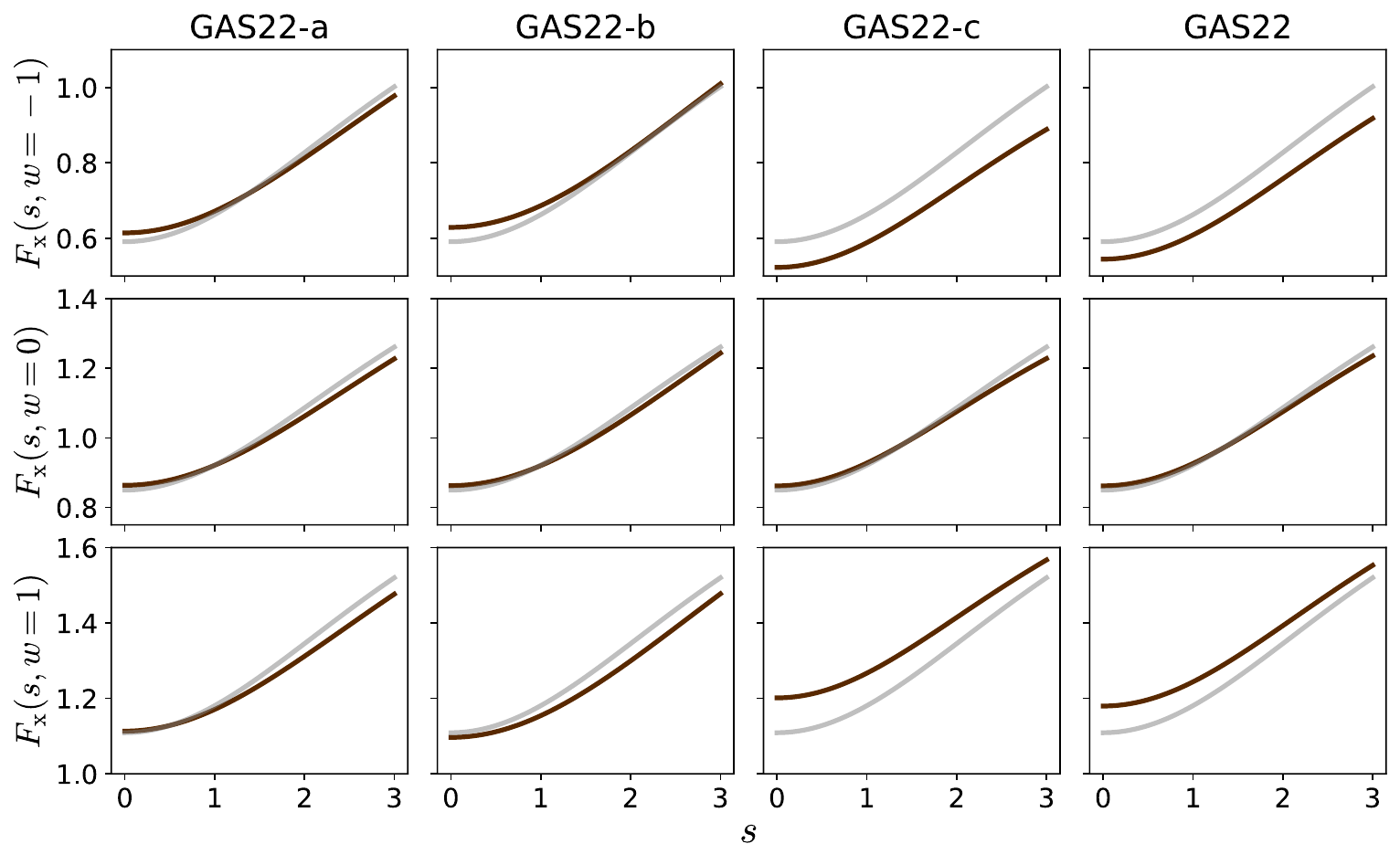}
  \caption{Exchange enhancement factors $F_\text{x}$ for functional forms in main text Fig. 4. For reference, the enhancement factor for the $\omega$B97M-V functional is plotted in grey.}
\end{figure*}

\begin{figure*}[!h]
  \centering
  \includegraphics[width=5in]{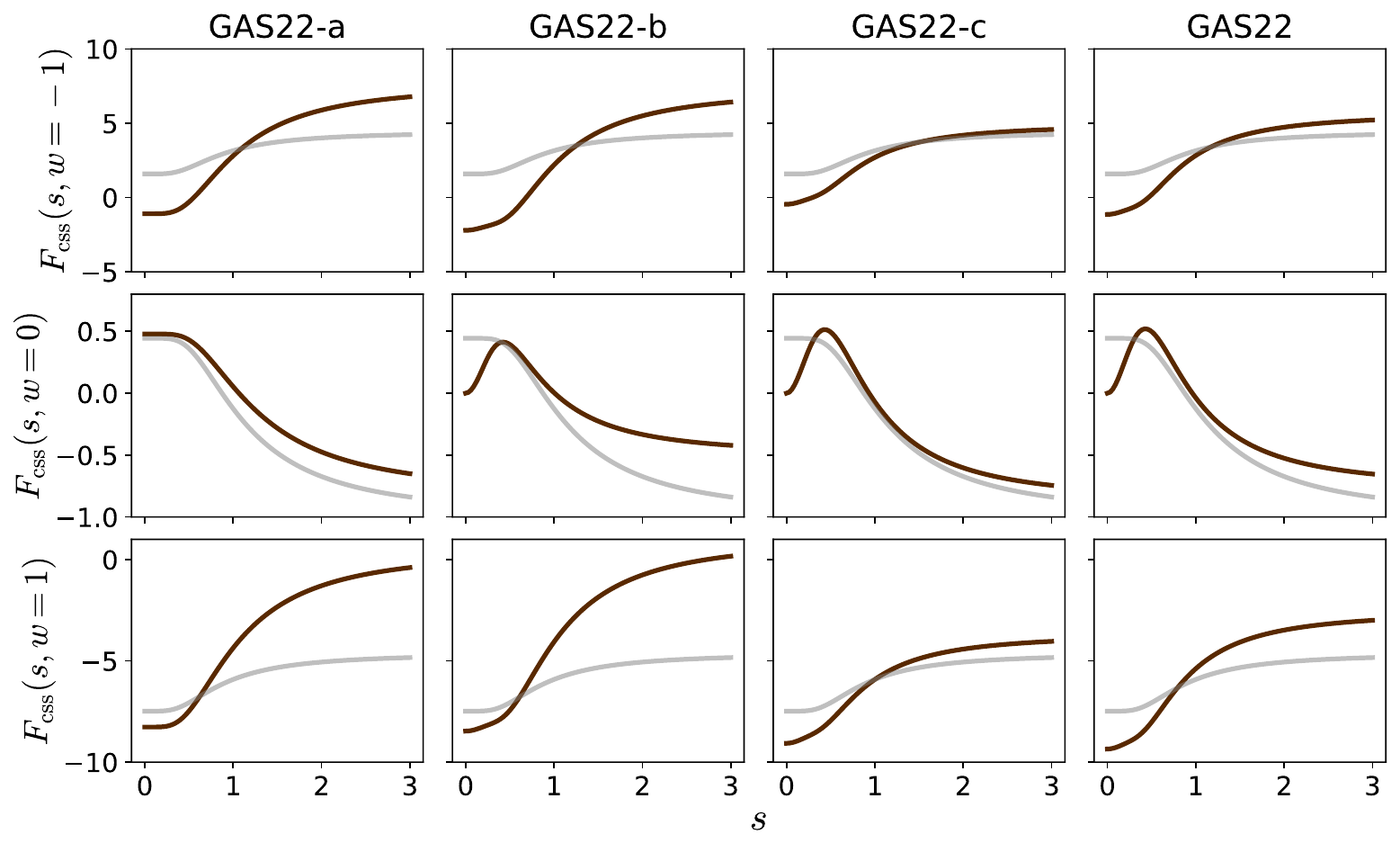}
  \caption{Same-spin correlation enhancement factors $F_\text{c-ss}$ for functional forms in main text Fig. 4. For reference, the enhancement factor for the $\omega$B97M-V functional is plotted in grey.}
\end{figure*}

\begin{figure*}[!h]
  \centering
  \includegraphics[width=5in]{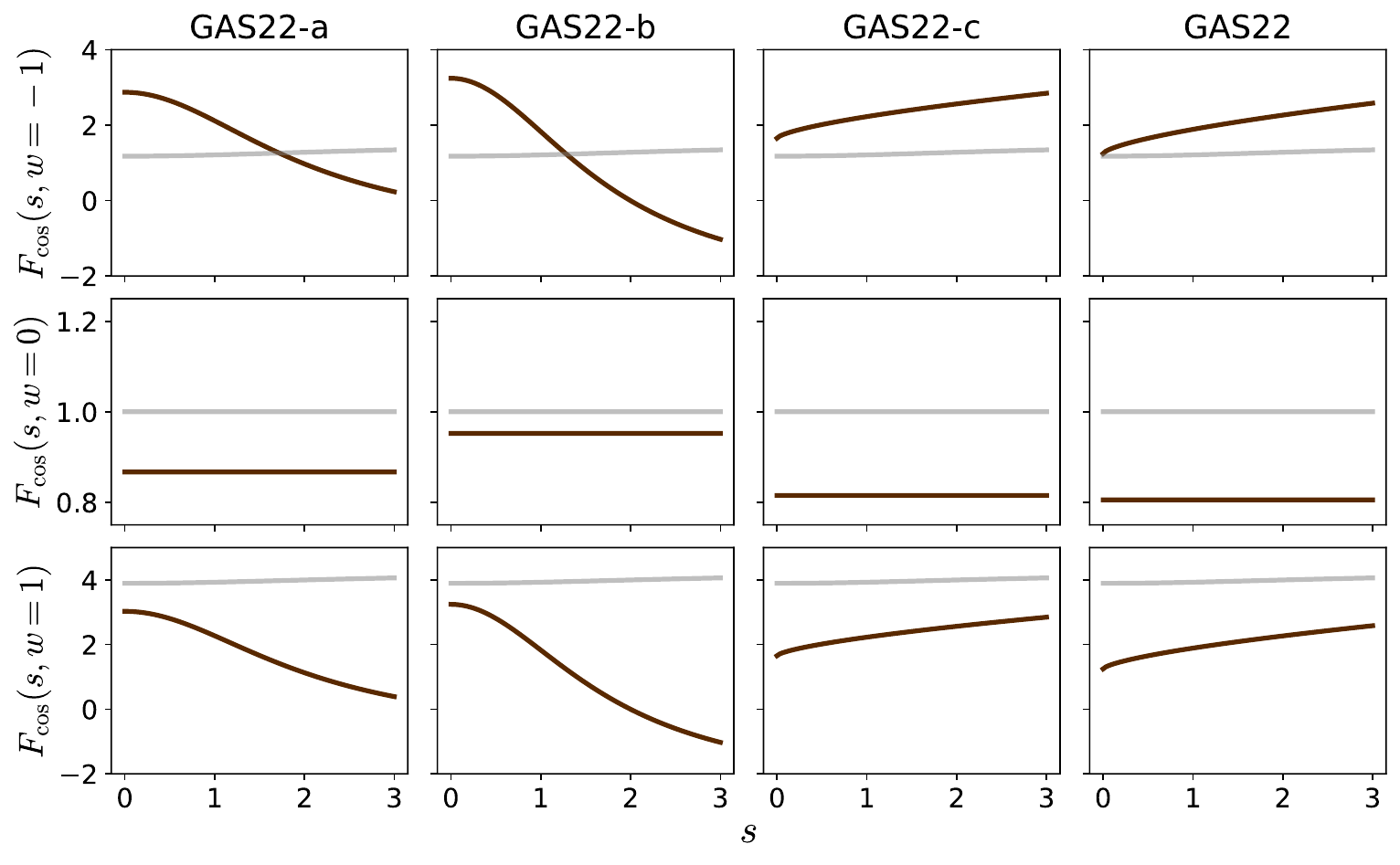}
  \caption{Opposite-spin correlation enhancement factors $F_\text{c-os}$ for functional forms in main text Fig. 4. For reference, the enhancement factor for the $\omega$B97M-V functional is plotted in grey.}
\end{figure*}

\pagebreak

\section{Random search studies starting from $\omega$B97M-V}

\begin{figure}[!h]
  \centering
  \includegraphics[width=4in]{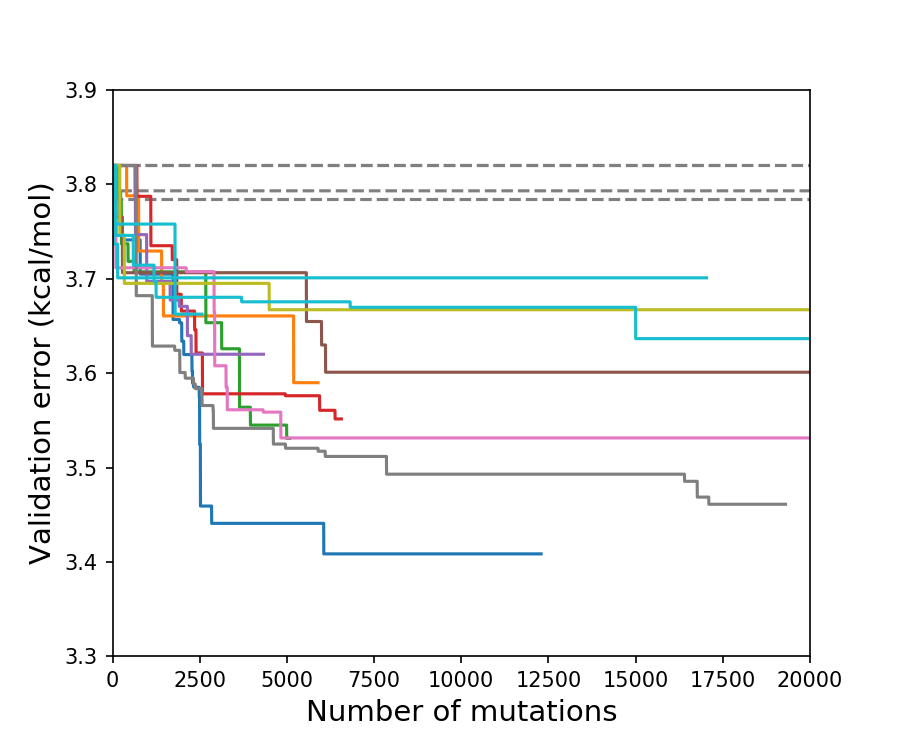}
  \caption{Validation error of symbolic functionals generated by random search and regularized evolution experiments starting from the $\omega$B97M-V functional. Random search and regularized evolution results are shown with dashed (solid) lines, with different lines represent independent experiments. The reference values in MGCDB84 database are used as targets for training and evaluation of functionals.}
  \label{wb97mv}
\end{figure}

In main text Fig. 3 we presented regularized evolution calculations starting from the $\omega$B97M-V functional. For comparison, in Fig. \ref{wb97mv} we report random search calculations (dash lines). The random search studies are performed with identical set up as regularized evolution experiments, except that the tournament size is set to 1. Therefore, in each iteration of random search experiment, the parent functional used for mutation is randomly selected from the population without referring to the fitness of functional forms. Compared to regularized evolution calculations, random search is found to be ineffective in traversing the search space and generating better functional forms than existing forms.

Here we make some additional remark on the starting point (termed GAS22-a in the main text) of the regularized evolution and random search studies. GAS22-a has identical symbolic form as $\omega$B97M-V, but with all parameters (including $\gamma$'s) optimized on the training set as done for all the symbolic functional forms generated in this work. In the original work that created the $\omega$B97M-V functional, the nonlinear parameters $\gamma$'s are not optimized and only linear parameters are optimized. Thus GAS22-a is a different functional as $\omega$B97M-V and have different training/validation/test errors: 2.97/3.82/4.47 kcal/mol.

\pagebreak

\section{Software design}

\begin{figure*}[!h]
  \centering
  \includegraphics[width=5in]{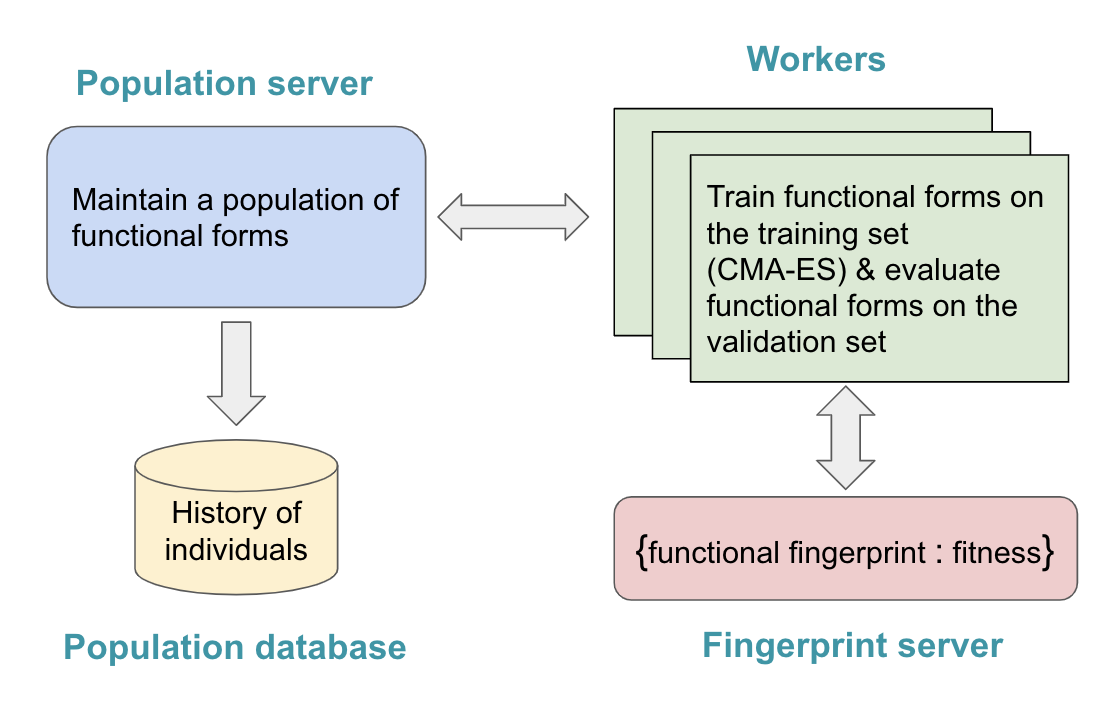}
  \caption{Design of symbolic regression software program. The program consists of a population server, a population database, a fingerprint server for functional equivalence checking and a number of workers for training and evaluating functional forms. The regularized evolution process is performed on the population server, and all child functionals are sent to workers for training and evaluation. The workers will first check if equivalence forms are already explored. If equivalence forms are explored before, the worker will directly send the cached fitness value in fingerprint server to the population server. }
  \label{system_design}
\end{figure*}

In Fig.~\ref{system_design} we present the high-level software design of the distributed regularized evolution program. The program consists of a population server, a population database, a fingerprint server for functional equivalence checking and a number of workers for training and evaluating functional forms. The training of a functional form is performed with the CMA-ES algorithm, which require to compute the the training error on tens of thousands of sets of different parameters. Such calculations are efficiently performed by porting the calculation of training errors to GPU processors through just-in-time compilation.

As briefly mentioned in the main text, one functional form may have multiple equivalent symbolic representations. For the purpose the functional equivalence checking, we define equivalent forms as forms that evaluates to the same value given same parameters and features, and we do not consider more complicated forms of equivalence such as the those requiring a mapping of parameters (e.g. the equivalence of B97 exchange functional and the symbolic functional obtained in main text).

To check for equivalent functional forms and avoid duplicated computations, each functional is assigned a \textit{fingerprint}. The fingerprint is evaluated by computing the functional values using a set of features and parameters that are randomly chosen but kept consistent during the entire program. The functional values are then hashed and the hash value serves as the functional fingerprint. The fingerprint is identical across all equivalent functional forms because they all evaluates to the same values with same parameters and features. All fingerprints and fitness values of explored functionals are cached during the regularized evolution calculations. Every time a new functional form is generated from mutation, its fingerprint will be evaluated to check if equivalent forms have already been explored. If equivalent forms are explored before, the cached fitness values are used without re-training the functional form.

\bibliography{references}

\bibliographystyle{Science}

\end{document}